\documentclass[letterpaper, 10 pt, conference]{ieeeconf}  

\IEEEoverridecommandlockouts                              

\newcommand{\acceptable}{\cellcolor[HTML]{90EE90}}

\title{\LARGE \bf
Human Supervisor Workload Prediction: Lag Horizon Selection}

\author{Mark-Robin Giolando$^{1}$ and Julie A. Adams, \textit{Senior Member, IEEE}$^{1}$
\thanks{$^{1}$Mark-Robin Giolando and Julie A. Adam are with the Collaborative Robotics and Intelligent Systems Institute at Oregon State University, Corvallis, OR, USA {\tt\small giolandm@oregonstate.edu, julie.a.adams@oregonstate.edu}}%
\thanks{The presented work was partially supported by ONR grants N00024-20-F-8705 and N00014-21-1-2052. The views, opinions, and findings expressed are those of the authors and are not to be interpreted as representing the official views or policies of the DOD or the U.S. Government.}%
\thanks{© 2026 IEEE.  Personal use of this material is permitted.  Permission from IEEE must be obtained for all other uses, in any current or future media, including reprinting/republishing this material for advertising or promotional purposes, creating new collective works, for resale or redistribution to servers or lists, or reuse of any copyrighted component of this work in other works}
}

\usepackage{cite}
\usepackage{amsmath,amssymb,amsfonts}
\usepackage{algorithmic}
\usepackage{graphicx}
\usepackage{textcomp}
\usepackage{xcolor}
\def\BibTeX{{\rm B\kern-.05em{\sc i\kern-.025em b}\kern-.08em
    T\kern-.1667em\lower.7ex\hbox{E}\kern-.125emX}}

\usepackage{subcaption}
\usepackage{tabu}
\usepackage{longtable}
\usepackage{multirow}
\usepackage{makecell}
\usepackage[table]{xcolor}
\usepackage[absolute,overlay]{textpos}

\begin{document}

\maketitle
\thispagestyle{empty}
\pagestyle{empty}

\begin{abstract}

Teleoperation systems must be aware of the human's workload during missions to maintain operator performance. Prior work employed wearable physiological sensor response metrics to estimate current human workload; however, these estimates only enable robots to respond to under- or overload conditions reactively. Current human workload prediction approaches are limited to very short prediction horizons and fail to investigate variable lag horizons' impact on those predictions. This manuscript investigates physiological sensor driven human workload prediction focusing on the impact of lag horizons on both univariate and multivariate time series forecasting models, with longer prediction horizons than the workload prediction state-of-the-art (i.e., $>$ 30 seconds using Long Short-Term Memory networks). Models were trained using data from a 64 participant non-sedentary supervisory environment NASA Multi-Attribute Task Battery-II human subjects evaluation. A key finding is that univariate workload predictions required 240 second lag horizons, whereas multivariate workload predictions sufficed with 120 second lag horizons. This finding indicates additional workload components reduce lag horizon requirements, enabling more efficient models with longer prediction horizons. 

\end{abstract}



\section{Introduction}





Teleoperating robots place multifaceted demands on the human, requiring resources from multiple workload channels. 
Workload represents the ratio between the resources a system's tasks demand, and the human's available resources to meet this demand \cite{yeh1988dissociation}. 
Operator performance may be negatively impacted by adverse workload conditions caused by task demands exceeding the human's available resources in an \textit{overload} condition, whereas insufficient task demand may lead to human disengagement (i.e., \textit{underload}) \cite{wickens2004introduction}. 
Robot teleoperation systems must account for the human's comprehensive overall workload (i.e., the cognitive, auditory, speech, visual, gross motor, fine motor, and tactile workload components) to address these suboptimal workload states. 
Prior work used wearable physiological sensor data to estimate the human's current workload for each workload component (i.e., cognitive, auditory, speech, visual, gross motor, fine motor, tactile) and the overall workload \cite{smith2022decomposed, bs2023cogsima, joshthesis}. 
However, these workload estimates only enable reactive actions to past workload states, after performance has already decreased. Predictive workload models of future workload states are necessary to preempt negative workload states before they occur by allowing time for a teleoperation system to plan and implement adaptations to maintain performance, where longer predictions (i.e., 1-4 minutes) enable more complex and effective adaptations.

The duration of historical data used for future predictions is the lag horizon, and the prediction horizon is the future time point predicted. 
Current workload prediction methods only focus on cognitive workload \cite{brand2017model, boehm2021real, pang2023air,wei2023classification, qin2023electroencephalogram, grimaldi2024deep}. 
Up to 30 second (s) workload predictions were made using prior workload values \cite{yu2023towards, grimaldi2024deep}, historical task loads \cite{yu2023towards, pang2023air}, physiological response metrics \cite{wei2023classification, qin2023electroencephalogram, grimaldi2024deep}, and performance metrics \cite{boehm2021real} as model predictors. 
However, these methods only predict a single workload component, have very short prediction horizons (i.e., generally $<5$s), and do not investigate the impact of varying lag horizons. 
Deciding the lag horizon remains an open question \cite{brand2017model}.

Longer lag horizons may result in more accurate and longer duration predictions; however, a minimum length lag horizon is desired to enable more computationally efficient models, and to avoid overtraining on the dataset. 
Prediction accuracy may be influenced by time series data trends (i.e., long term fluctuations), and seasonality (i.e., fixed patterns or cycles) \cite{hyndman2018forecastingCh3}. Autocorrelation lengths (i.e., the similarity between a time series and a lagged version of itself) may also determine how long a data point influences future data points, impacting the maximum useful lag horizon \cite{flores2012autocorrelation, kandananond2012comparison}.

Lag horizons are often overlooked due to the main focus being on prediction accuracy and horizon length; however, lag horizon has a significant impact on the predictive output 
(e.g., longer lag horizons increasing accuracy and prediction horizon, while also increasing compute demand). 
This manuscript investigated minimum and maximum lag horizons for cognitive, visual, auditory, and overall workload predictions using three prediction horizons for a human operating a simulated aerial robot. These prediction models will enable future teleoperated robots and interfaces to preemptively adapt to the human's workload levels.

\section{Background}

Preemptive adaptation necessitates workload prediction algorithms as workload estimates are unable to provide accurate predictions of the human's future workload. Thus, workload estimation algorithms can only retroactively react to adverse workload conditions, which when coupled with the communication delays can lead to adaptation thrashing. 

Long-term workload predictions (i.e, 1-24 hours) may be used for pre-mission planning to confirm participants will have sufficient resources for the task demands \cite{abreu2023data}. 
Mission plans were created to normalize the human's workload, using the expected duration and workload levels of each task \cite{van2017use, fas2019ntom, fas2019dynamic}. 
These methods are generally unresponsive to environmental dynamics, which may result in poor predictions and ineffective adaptations \cite{cullen1999validation} resulting in additional demands placed on the human and decreased mission performance.

Short-term predictions make predictions mid-mission to account for dynamic missions \cite{boehm2021real, pang2023air, yu2023towards, qin2023electroencephalogram, wei2023classification, grimaldi2024deep}. 
These methods have used prior workload values \cite{yu2023towards, grimaldi2024deep}, historical task loads \cite{yu2023towards, pang2023air}, physiological response metrics \cite{grimaldi2024deep}, and performance metrics \cite{boehm2021real} as model predictors.  These predictors were incorporated into linear regression \cite{yu2023towards}, long short-term memory networks \cite{grimaldi2024deep}, transformers \cite{wei2023classification, grimaldi2024deep}, and two-part \cite{boehm2021real} models to predict cognitive workload. 
These approaches generally made predictions $\leq$ 5s \cite{boehm2021real, pang2023air, yu2023towards, qin2023electroencephalogram, wei2023classification}, with one approach predicting up to 30s \cite{grimaldi2024deep}. 
These predictions are able to quickly react to changes in the human's workload; however, these predictions generally only predict the workload class (i.e., overload, normal load, underload), and do not predict the capacity on the remaining six components. 
These limitations provide insufficiently detailed workload predictions to properly inform adaptations, and may not provide sufficient time for adaptation to occur. 
Furthermore, these workload algorithms lack validation against baseline methods to confirm that future workload predictions are actually made, rather than just current workload estimates.


Workload may be treated as a time series \cite{levin2006tracking, rusnock2015objective}. 
Prior and current workload data points may be used to predict the distribution of future observations (i.e., p($y_{t+h} | y_{1:t}$,$u_{1:t}$), where $y_{t}$ is the workload time series, $h$ is the prediction horizon length, and $u_t$ refers to input covariates). 
The historical data durations are \textit{lag horizons} and the future time durations are the \textit{prediction horizons}. Longer prediction horizons enable a greater variety of adaptations. Shorter lag horizons enable more efficient models, while longer lag horizons generally increase prediction accuracy up to the point where out-of-date data is included. 

\section{Methods}
 

A NASA Multi-Attribute Task Battery-II (MATB) \cite{santiago2011multi} evaluation simulated a supervisory-based human-robot team. 
The mixed-design evaluation manipulated tasks, task density, and workload ordering. 
Task density manipulated the number of tasks initiated during a specific time period by 
varying the tasks’ frequency in three levels, each corresponding to a relative workload level: underload (UL), normal load (NL), and overload (OL) \cite{weinger1994objective}. The workload ordering variable ensured each workload transition occurred exactly once. 64 participants (37 male, 24 female, and 3 non-binary) ages 18-60 completed a single 52.5 minute (min) trial of seven consecutive 7.5 min task density conditions 
(i.e., UL-NL-OL-UL-OL-NL-UL, NL-OL-UL-OL-NL-UL-NL, OL-UL-OL-NL-UL-NL-OL). 
Participants completed a consent form upon arrival.


\begin{table}
\centering
\caption{The wearable sensors and multi-dimensional workload estimation algorithm metrics for each component.}

\resizebox{\columnwidth}{!}{%
\begin{tabular}{|c|c|}
\hline
\textbf{Sensor} & \textbf{Metric}  \\ \hline
 & Heart Rate  \\ \cline{2-2} 
 & Heart Rate Variability    \\ \cline{2-2} 
 & Respiration Rate  \\ \cline{2-2} 
\multirow{-4}{*}{Bioharness} & Postural Magnitude  \\ \hline
 & Pupil Diameter  \\ \cline{2-2} 
 & Blink Latency  \\ \cline{2-2} 
 & Blink Rate   \\ \cline{2-2} 
 & Index of Pupillary Activity  \\  \cline{2-2} 
 & Fixations  \\ \cline{2-2} 
 & Gaze Entropy  \\ \cline{2-2} 
\multirow{-7}{*}{Pupil Core} & Saccades  \\ \hline
\end{tabular}

\begin{tabular}{|c|c|}
\hline
\textbf{Sensor} & \textbf{Metric}  \\ \hline
 & Voice Intensity  \\ \cline{2-2} 
 & Voice Pitch  \\ \cline{2-2} 
 & Speech rate   \\ \cline{2-2} 
 & Mel-Frequency  \\ 
 &  Cepstral Coefficients    \\ \cline{2-2} 
\multirow{-5}{*}{Microphone} & Spectrogram  \\ \hline
\begin{tabular}[c]{@{}c@{}}Reed\\ Decibel Meter\end{tabular} & Noise level   \\ \hline
Xsens & Inertial \\ \hline
 & Inertial \\ \cline{2-2} 
\multirow{-2}{*}{Myo armband} & Surface Electromyography   \\ \hline
\end{tabular}
}

\label{tab:metrics_selection}
\end{table}

Physiological response metrics were collected 
by a BioPac Bioharness BT, Xsens Mtw Awinda motion trackers, a Pupil Lab's Core eye tracker, two Myo devices, a noise meter, and a Shure Microphone to collect physiological metrics for workload estimation, as provided in Table \ref{tab:metrics_selection}. 
The cognitive component used heart rate, heart rate variability, pupil diameter, blink latency and noise level, while the visual component depended on pupil diameter, blink latency, blink rate, IPA, fixations, gaze entropy and saccades. 
The speech component used voice intensity, voice pitch, speech rate, Mel-Frequency Cepstral Coefficients (MFCC), and spectrogram, whereas the auditory relied on MFCC, spectrogram and noise level. 
The gross motor metrics were heart rate, respiration rate, postural magnitude and Xsens inertial metrics, the fine motor used Xsens and Myo inertial metrics as well as surface electromyography, and the tactile component metrics were Xsens inertial metrics and surface electromyography.


The supervisory task environment was a modified MATB \cite{santiago2011multi}, where the operator supervised a simulated remotely-piloted aerial vehicle while managing four tasks: tracking, system monitoring, resource management, and communication. The communication task was decomposed into two tasks (i.e., communication and communication response). Modification of the MATB physically separated each task, requiring participants to walk between the two stations. 
The required equipment (e.g., joystick or a keyboard) to complete each task was placed in front of the respective monitor.


\begin{figure}
    \centering
    \begin{subfigure}[b]{0.23\textwidth}
        \centering
        \includegraphics[height = 3.75cm, width = \textwidth]{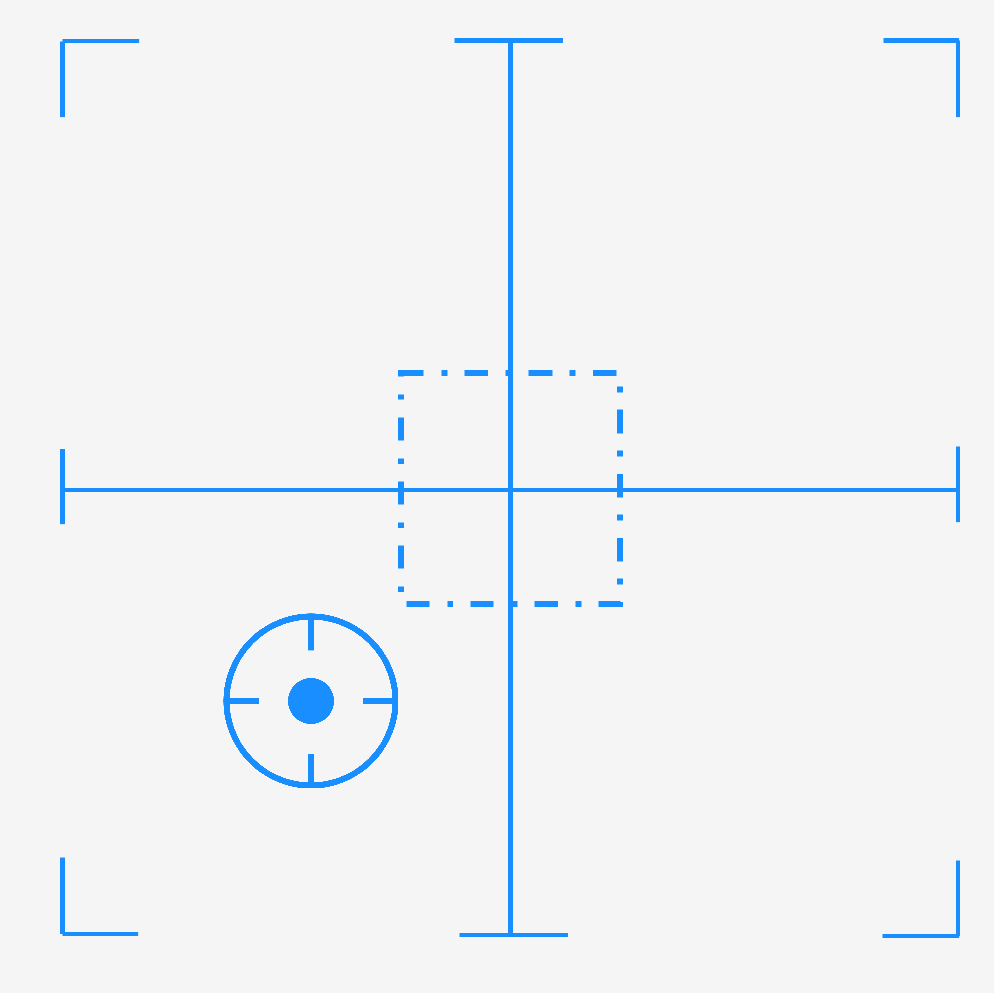}
        \caption{\small Tracking}
        \label{fig:tracking_task}
    \end{subfigure}
    \begin{subfigure}[b]{0.23\textwidth}  
        \centering 
        \includegraphics[height = 3.75cm, width = \textwidth]{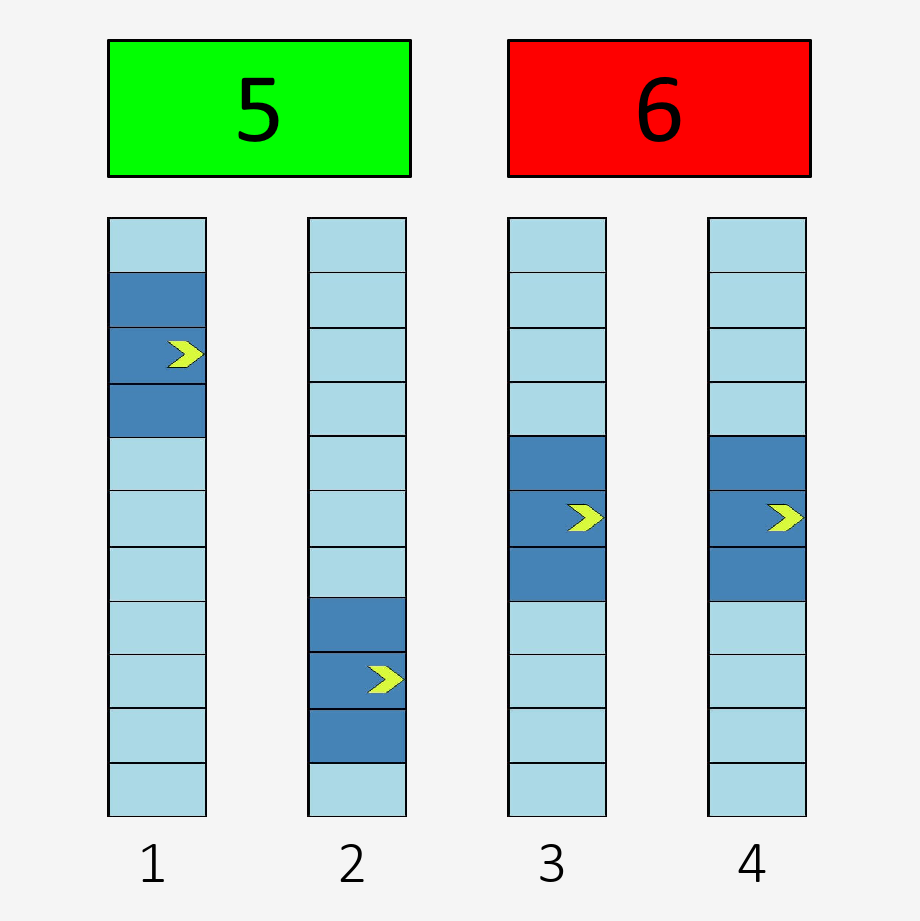}
        \caption{\small System Monitoring}
        \label{fig:sysmon_task}
    \end{subfigure}
    \vskip\baselineskip
    \begin{subfigure}[b]{0.23\textwidth}   
        \centering 
        \includegraphics[height = 3.75cm, width = \textwidth]{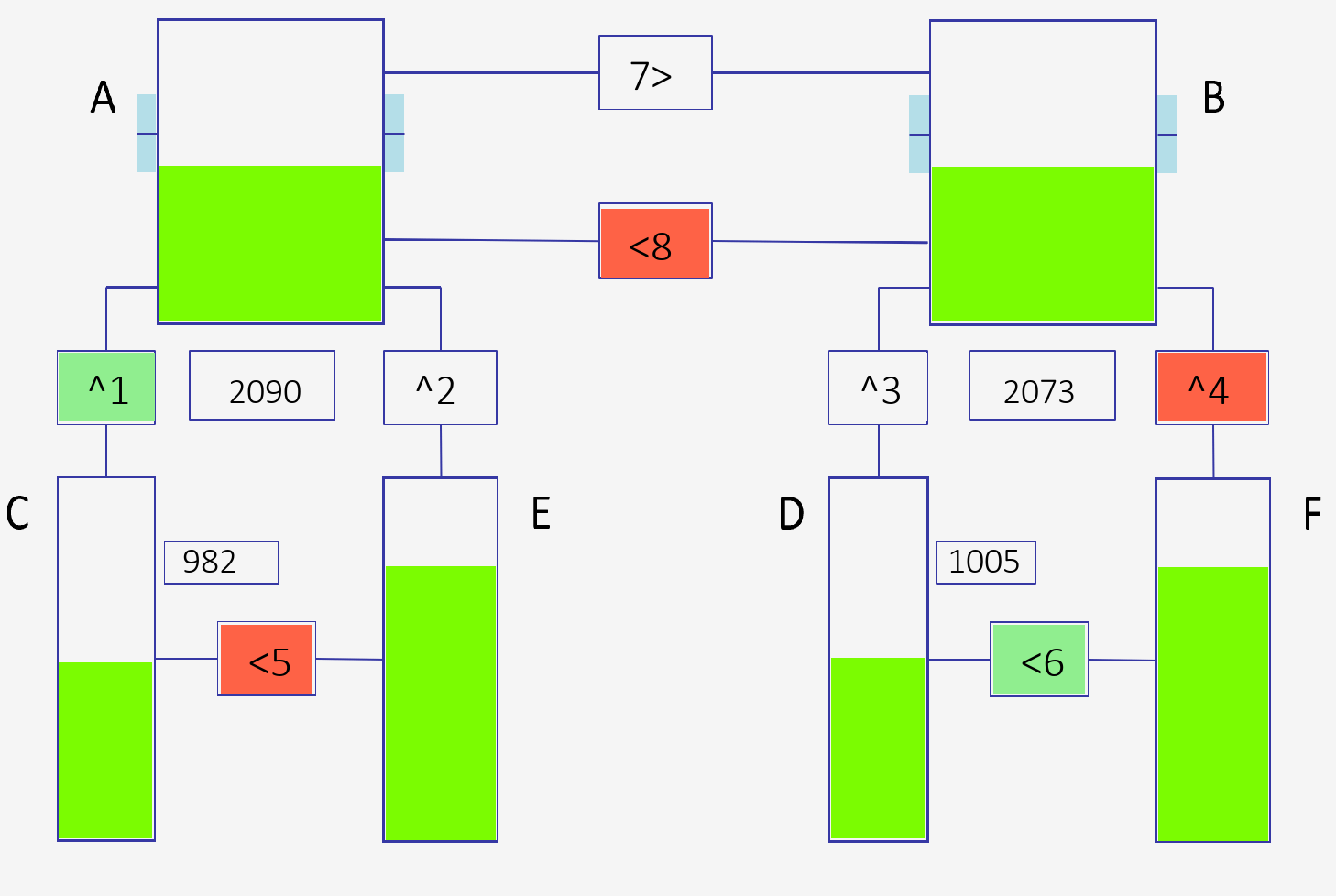}
        \caption{\small Resource Management}
        \label{fig:resman_task}
    \end{subfigure}
    \begin{subfigure}[b]{0.23\textwidth}   
        \centering 
        \includegraphics[height = 3.75cm, width = \textwidth]{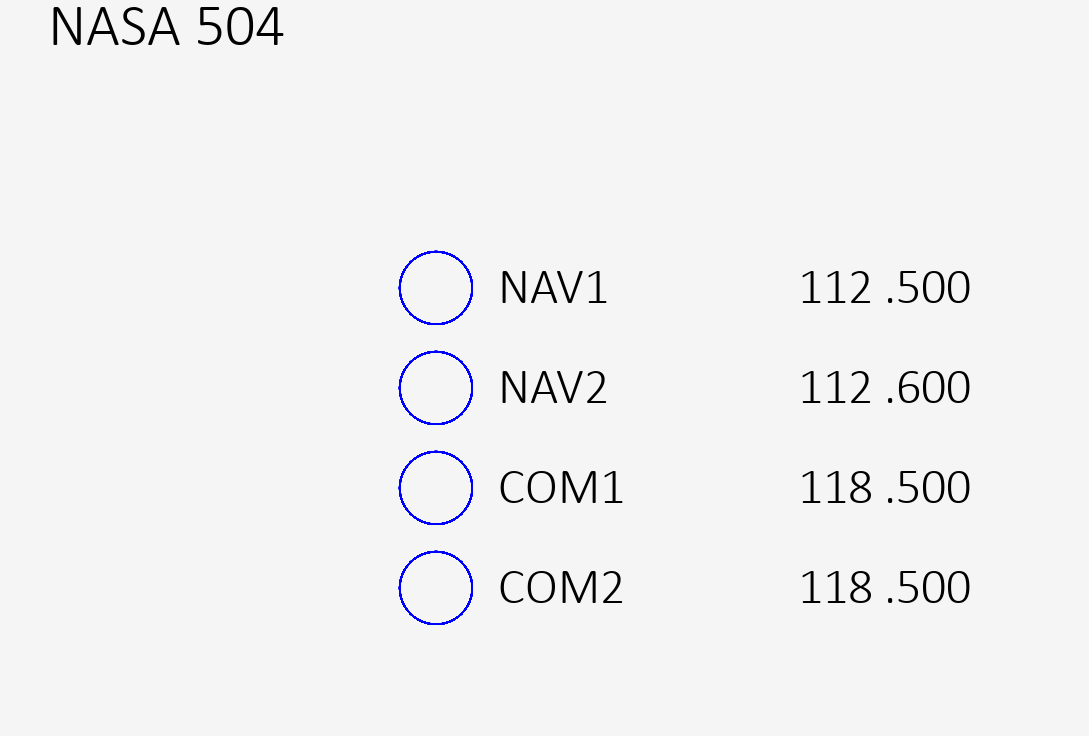}
        \caption{\small Communications}
        \label{fig:comm_task}
    \end{subfigure}
    \caption{\small The NASA MATB-II Tasks.}
    \label{fig:matb_task}
\end{figure}

The tracking task, depicted in Fig. \ref{fig:tracking_task},
required participants to keep the circle in the middle of the cross-hairs using a joystick. 
The system monitoring task, shown in Fig. \ref{fig:sysmon_task}, 
required monitoring two colored lights and four gauges. If the green or the red light turned on, an out of range value required resetting. The four gauge indicators randomly moved up and down, typically remaining in the middle. Resetting required pressing the corresponding number key on the keyboard. 
The resource management task included six fuel tanks (A-F) and eight fuel pumps (1-8), shown in Fig. \ref{fig:resman_task}
. The arrow by the fuel pump’s number indicated the fuel flow direction. Participants were to maintain the fuel levels of Tanks A and B by turning the fuel pumps on or off. Fuel Tanks C and D had finite fuel levels, while Tanks E and F had an infinite supply. A pump turned red when it was unable to pump fuel. 
The communications task required listening to air-traffic control requests for radio changes, ignoring requests not directed to the participant's call sign. 
Participants changed the specified radio to the specified frequency by selecting the desired radio and using arrows to change the radio’s frequency, as depicted in Fig. \ref{fig:comm_task},
and provided a verbal response. 
Participants walked around the tables to attend to the other tasks whenever they heard a ping sound. Participants were free to move between tasks at any time, but the ping enforced a mandatory transition.

The tracking task had one 45s manual instance in the UL condition, one 20s manual session per minute in the NL condition, and two 12s manual sessions per minute in the OL condition. 
The system monitoring task had one instance in the UL condition, five instances per minute in the NL condition and 15 instances per minute in the OL condition. 
The resource management task had two minutes of manual control with no pump failures in the UL condition, 3.5 minutes with two pump failures per minute in the NL condition and was entirely manual with two or more pump failures per minute in the OL condition. 
The communications task had one request and response in UL, two requests and one response per minute in NL, and three requests with two or more responses per minute in the OL condition.

Task timings and occurrences were chosen such that the correct workload condition, or task density, was elicited. 
The IMPRINT Pro human performance modeling tool was used to model the tasks for each workload level and ordering prior to conducting the evaluation and provided anchors for selecting a task's correct \textit{workload} value \cite{mitchell2000mental}, 
as well as to provide continuous workload values to act as pseudo ground truth labels for all workload machine learning models.

\subsection{Prediction Model Design and Validation}

Two direct time series forecasting methods were evaluated, with historical workload estimates provided via pretrained workload estimation models \cite{smith2022decomposed, bs2023cogsima, joshthesis}. 
The univariate approach used only the overall historical workload estimates, or that of a single workload component, to predict the same component's future workload. 
The multivariate approach used all seven components' estimated workload to predict future workload. 
Both approaches used autoregressive feed forward neural networks with a ``global" dataset of similar time series (i.e., other humans executing the mission plan) \cite{MurphyCh29}, and a leave-one-subject-out five-fold blocked cross validation method \cite{cerqueira2020evaluating}. 
The cross validation divided the data by timestamp, iteratively using a single block from one participant for test data, and the remaining four blocks from the other participants for training data. 
Models were trained with data from each participant with sufficient data to use the blocked cross validation (e.g., $>$ 30 minutes for 120 second (s) predictions using 240s lag horizons). 
Each four-layer neural network was implemented using PyTorch 2.2.0, NumPy 1.26.3 and SciPy 1.11.4 with two hidden layers of 128 nodes each. Models were trained using the Adam optimizer with a learning rate of 0.0001, a 128 batch size, and used early stopping to avoid overfitting. 
The models employed a 5s step size using prediction horizons of 60s, 120s, and 240s, and lag horizons of 30s, 60s, 120s and 240s. 
The prediction horizons and lag horizons were chosen to evaluate the maximum range afforded by the five folds (i.e., 600s). 
The mean cognitive, visual, auditory and overall autocorrelation lengths were identified by autocorrelation functions to be 135, 134, 124 and 137, respectively.


\section{Results}


Spearman correlations provide analysis of the similarity between the univariate or multivariate predictions and IMPRINT Pro, and were used to evaluate the models' performance. 
The Benjamini–Hochberg procedure was used for multi-comparison correction. 
Prior work established correlation values of $>0.7$ are considered high correlation, 0.5-0.69 are moderate, whereas correlation values below 0.5 are low, with values $<0.3$ regarded as negligible \cite{hinkle2003applied}. 
Moderate correlation values are deemed acceptable for prediction, with high correlation values preferred. 
The MATB environment primarily induced cognitive and visual workload. 
Cognitive, visual, auditory and overall workload predictions are presented using both univariate and multivariate methods, alongside minimum and maximum lag horizon analysis. 
The remaining workload components are excluded due to insufficient lag horizon differences, or low correlation values.

The univariate and multivariate models were validated against a na\"ive persistence baseline model, shown in Tables \ref{tab:Univariate_corr} and \ref{tab:multivariate_corr}. Both approaches significantly outperformed the persistence model for all workload components, lag horizons and prediction horizons (p $<$ 0.05). 
Green cells in each table indicate the mean correlation values are $>$ 0.5, and therefore represent acceptable lag-prediction horizon pairs.

\subsection{Univariate Workload Prediction}

\begin{table*}
\centering
\caption{The univariate cognitive, visual, auditory and overall workload prediction Spearman correlations [mean (std)] by prediction and lag horizon. Green cells indicate acceptable correlation. }

    \begin{tabular}{|c|c|c|c|c|c|} \hline
    
 \multicolumn{6}{|c|}{\textbf{Cognitive}} \\ \hline
\begin{tabular}[c]{@{}l@{}}\textbf{Pred.}\\ \textbf{Hor.}\end{tabular}  & \textbf{Na\"ive} & \textbf{30s Lag}  & \textbf{60s Lag} & \textbf{120s Lag} & \textbf{240s Lag} \\ \hline
 \textbf{60s}  & 0.45 (0.13)  & 0.48 (0.14)   & \acceptable0.52 (0.15)  & \acceptable0.56 (0.14)   & \acceptable0.64 (0.12)  \\ \hline
 \textbf{120s} & 0.42 (0.14) & 0.40 (0.13)   & 0.41 (0.13)  & 0.44 (0.13)   & \acceptable0.54 (0.13)  \\ \hline
 \textbf{240s} & 0.12 (0.14) & 0.25 (0.08)   & 0.25 (0.09)  & 0.31 (0.08)   & \acceptable0.57 (0.19)  \\  \hline

 \multicolumn{6}{|c|}{\textbf{Visual}} \\ \hline
  \textbf{60s} &  0.44 (0.13) &  0.46 (0.13)  & 0.48 (0.13)  & 0.49 (0.13)   & \acceptable0.63 (0.11)  \\ \hline
 \textbf{120s} & 0.32 (0.10)  &  0.35 (0.11)  & 0.37 (0.11)  & 0.39 (0.11)   & \acceptable0.54 (0.11)  \\ \hline
 \textbf{240s} & 0.09 (0.14) &  0.25 (0.11)  & 0.25 (0.11)  & 0.27 (0.09)   & \acceptable0.57 (0.16)  \\  \hline

 \multicolumn{6}{|c|}{\textbf{Auditory}} \\ \hline
 \textbf{60s}  & 0.20 (0.18) & 0.35 (0.12)   & 0.48 (0.18)  & \acceptable0.57 (0.21)   & \acceptable0.63 (0.13)  \\ \hline
 \textbf{120s} & 0.32 (0.25) & 0.41 (0.15)   & 0.46 (0.15)  & \acceptable0.53 (0.18)   & \acceptable0.59 (0.12)  \\ \hline
 \textbf{240s} & 0.17 (0.19) & 0.29 (0.09)   & 0.32 (0.09)  & 0.41 (0.11)   & \acceptable0.57 (0.13)  \\  \hline

 \multicolumn{6}{|c|}{\textbf{Overall}} \\ \hline
 \textbf{60s}  & \acceptable0.54 (0.11) & \acceptable0.56 (0.12)   & \acceptable0.57 (0.13)  & \acceptable0.59 (0.13)   & \acceptable0.68 (0.09)  \\ \hline
 \textbf{120s} & 0.40 (0.12) & 0.43 (0.13)   & 0.43 (0.13)  & 0.46 (0.13)   & \acceptable0.59 (0.09)  \\ \hline
 \textbf{240s} & 0.14 (0.15) & 0.21 (0.09)   & 0.23 (0.10)  & 0.26 (0.09)   & \acceptable0.57 (0.14)  \\  \hline

\end{tabular}

\label{tab:Univariate_corr}
\end{table*}

The univariate cognitive, visual, auditory and overall workload predictions' Spearman correlations increased as the lag horizon increased for each prediction horizon, as shown in Table \ref{tab:Univariate_corr}. 
60s overall workload prediction horizons required a minimum 30s lag horizon to attain an acceptable correlation, while the 120s and 240s prediction horizons required 240s lag horizons. 
Overall workload predictions significantly differed between lag horizons per the Friedman test for the 60s (${\chi}^2(3,21) = 39.17$, p $<$ 0.001, \textit{W} = 0.49), 120s (${\chi}^2(3,20) = 42.26$, p $<$ 0.001, \textit{W} = 0.56) and 240s prediction horizons (${\chi}^2(3,20) = 42.36$, p $<$ 0.001, \textit{W} = 0.56). 
The Wilcoxon signed-ranked test indicated significant differences between lag horizons for each prediction horizon (p $<$ 0.05, 0.64 $<$ Cohen's \textit{d}).




The 60s univariate cognitive workload prediction horizons required a minimum 60s lag horizon for acceptable correlation and the 120s and 240s prediction horizons required 240s lag horizons. 
The cognitive workload prediction lag horizons differed significantly per the Friedman test for the 60s (${\chi}^2(3,37) = 65.30$, p $<$ 0.001, \textit{W} = 0.45), 120s (${\chi}^2(3,37) = 51.26 $, p $<$ 0.001, \textit{W} = 0.36) and 240s prediction horizons (${\chi}^2(3,37) = 63.26 $, p $<$ 0.001, \textit{W} = 0.44). 
The Wilcoxon signed-rank test indicated the univariate prediction Spearman correlations increased significantly with longer lag horizons for almost all prediction horizons, except the 30s and 60s lag horizons for the 240s prediction horizon (p $<$ 0.001, 0.86 $<$ Cohen's \textit{d}). 




Each visual workload prediction horizon required 240s lag horizons to achieve acceptable correlation, with the Friedman test identifying significant differences between lag horizons for the 60s (${\chi}^2(3,37) = 71.11$, p $<$ 0.001, \textit{W} = 0.49), 120s (${\chi}^2(3,37) = 76.23$, p $<$ 0.001, \textit{W} = 0.53) and 240s (${\chi}^2(3,37) = 52.75$, p $<$ 0.001, \textit{W} = 0.37) prediction horizons. 
The Wilcoxon signed-rank test indicated significant Spearman correlation improvements as lag horizon increased for the 60s and 120s prediction horizons (p $<$ 0.01, 0.77 $<$ Cohen's \textit{d}). The only significant difference for the 240s prediction horizon was observed between the 120s and 240s lag horizons (p $<$ 0.001, 27.11 $<$ Cohen's \textit{d}).




Auditory workload's 60s and 120s prediction horizons required a minimum 120s lag horizon for acceptable correlation. Acceptable results for the 240s prediction horizon required a 240s lag horizon. 
The Friedman test found significant differences 
between lag horizons for the 60s (${\chi}^2(3,52) = 118.96$, p $<$ 0.001, \textit{W} = 0.58), 120s (${\chi}^2(3,52) = 109.27$, p $<$ 0.001, \textit{W} = 0.54), and 240s (${\chi}^2(3,52) = 112.73$, p $<$ 0.001, \textit{W} = 0.55) prediction horizons. 
The Wilcoxon signed-rank test showed significant Spearman correlation differences between all lag horizons for all prediction horizons (p $<$ 0.01, 1.49 $<$ Cohen's \textit{d}). 








\subsection{Multivariate Workload Prediction}

\begin{table*}
\centering

\caption{The multivariate cognitive, visual, auditory and overall workload prediction Spearman correlations [mean (std)] by prediction and lag horizon. Green cells indicate acceptable correlation. }
    \begin{tabular}{|c|c|c|c|c|c|} \hline
 \multicolumn{6}{|c|}{\textbf{Cognitive}} \\ \hline
\begin{tabular}[c]{@{}l@{}}\textbf{Pred.}\\ \textbf{Hor.}\end{tabular}  & \textbf{Na\"ive} & \textbf{30s Lag}  & \textbf{60s Lag} & \textbf{120s Lag} & \textbf{240s Lag} \\ \hline
 \textbf{60s}   &  0.45 (0.13) &  \acceptable0.62 (0.08)  & \acceptable0.64 (0.07)  & \acceptable0.68 (0.09)   & \acceptable0.70 (0.10)  \\ \hline
 \textbf{120s}  & 0.36 (0.14)  &  \acceptable0.51 (0.11)  & \acceptable0.55 (0.09)  & \acceptable0.56 (0.10)   & \acceptable0.53 (0.11)  \\ \hline
 \textbf{240s}  & 0.12 (0.14)  &  0.28 (0.11)  & 0.30 (0.11)  & 0.31 (0.12)   & \acceptable0.53 (0.21)  \\  \hline

 \multicolumn{6}{|c|}{\textbf{Visual}} \\ \hline
 \textbf{60s}    & 0.44 (0.13) & \acceptable0.61 (0.08)   & \acceptable0.63 (0.06)  & \acceptable0.64 (0.11)   & \acceptable0.71 (0.09)  \\ \hline
 \textbf{120s}   & 0.32 (0.10) & 0.45 (0.08)   & 0.48 (0.07)  & \acceptable0.50 (0.11)   & \acceptable0.57 (0.09)  \\ \hline
 \textbf{240s}   & 0.09 (0.14) & 0.29 (0.12)   & 0.28 (0.11)  & 0.33 (0.11)   & \acceptable0.62 (0.11)  \\  \hline

 \multicolumn{6}{|c|}{\textbf{Auditory}} \\ \hline
 \textbf{60s}    & 0.20 (0.18)  &  \acceptable0.55 (0.11)   & \acceptable0.60 (0.11)   & \acceptable0.63 (0.10)    & \acceptable0.67 (0.10)   \\ \hline
 \textbf{120s}   & 0.32 (0.25)  &  \acceptable0.52 (0.05)   & \acceptable0.53 (0.07)   & \acceptable0.55 (0.07)    & \acceptable0.62 (0.09)   \\ \hline
 \textbf{240s}   & 0.17 (0.19)  &  0.35 (0.06)   & 0.37 (0.09)   & 0.41 (0.08)    & \acceptable0.54 (0.11)   \\ \hline

 \multicolumn{6}{|c|}{\textbf{Overall}} \\ \hline
 \textbf{60s}    & \acceptable0.54 (0.11) &  \acceptable0.62 (0.08)  &  \acceptable0.65 (0.07) & \acceptable0.68 (0.10)   & \acceptable0.70 (0.05)  \\ \hline
 \textbf{120s}   & 0.40 (0.12) &  \acceptable0.51 (0.12)  &  \acceptable0.55 (0.11) & \acceptable0.57 (0.09)   & \acceptable0.58 (0.08)  \\ \hline
 \textbf{240s}   & 0.14 (0.15) &  0.28 (0.12)  &  0.30 (0.12) & 0.33 (0.13)   & \acceptable0.62 (0.12)  \\  \hline

\end{tabular}
\label{tab:multivariate_corr}
\end{table*}

Spearman correlation values for multivariate predictions of overall, cognitive, auditory and visual workload increased as the lag horizon increased for each prediction horizon, as shown in Table \ref{tab:multivariate_corr}. 
The 60s overall workload prediction horizons required a minimum 30s lag horizon for acceptable correlation, and a 240s lag horizon for high accuracy. 
The 120s and 240s prediction horizons required 30s and 240s lag horizons for acceptable correlation, respectively. 
The Friedman tests indicated significant differences in overall workload predictions between lag horizons for 60s (${\chi}^2(3,21) = 28.26$, p $<$ 0.001, \textit{W} = 0.35), 120s (${\chi}^2(3,20) = 25.68$, p $<$ 0.001, \textit{W} = 0.34) and 240s prediction horizons (${\chi}^2(3,20) = 36.90$, p $<$ 0.001, \textit{W} = 0.49). 
The 60s and 120s prediction horizon correlations significantly increased between the 30s, 60s and 120s lag horizons (p $<$ 0.05, 1.89 $<$ Cohen's \textit{d}) per the Wilcoxon signed-rank test. The 240s prediction horizon correlations only significantly increased between the 120s and 240s (p $<$ 0.001, 1.87 $<$ Cohen's \textit{d}) lag horizons. 



Cognitive workload's 60s predictions required a 30s lag horizon for acceptable correlation, and a 240s lag horizon for high correlation. Acceptable results for 120s and 240s prediction horizons required 30s and 240s lag horizons, respectively. 
The Friedman test indicated the significant Spearman correlation differences between lag horizons for the 60s (${\chi}^2(3,21)  = 27.91$, p $<$ 0.001, \textit{W} = 0.35), and 
120s (${\chi}^2(3,20) = 18.42$, p $<$ 0.001, \textit{W} = 0.24), 
240s (${\chi}^2(3,20) = 23.82$, p $<$ 0.005, \textit{W} = 0.31) prediction horizons. 
The Wilcoxon signed-rank test found significant increases between the 30s, 60s and 120s lag horizons within the 60s prediction horizon, and between the 30s vs.\ 60s and 120s vs.\ 240s lag horizons within the 60s and 240s prediction horizons, respectively (p $<$ 0.05, 2.04 $<$ Cohen's \textit{d}). 



The 60s visual workload prediction horizons required a 30s lag horizon for acceptable correlation, and a 240s lag horizon for high correlation. The 120s and 240s prediction horizons required 120s and 240s lag horizons, respectively, to achieve acceptable results. 
The Friedman test for the visual multivariate prediction indicated significant differences between lag horizons for the 60s (${\chi}^2(3,21) = 24.60 $, p $<$ 0.001, \textit{W} = 0.31),  120s (${\chi}^2(3,20) = 22.26 $, p $<$ 0.001, \textit{W} = 0.29) and 240s (${\chi}^2(3,20) = 41.58 $, p $<$ 0.001, \textit{W} = 0.55) prediction horizons. 
Wilcoxon signed-ranked tests showed significant increases in Spearman correlation as the lag horizon increased for the 30s, 60s and 120s horizons within the 60s and 120s prediction horizons based on the Wilcoxon signed-rank test (p $<$ 0.05, 5.17 $<$ Cohen's \textit{d}). 
Furthermore, the correlations significantly increased as the lag horizons increased for the 60s, 120s and 240s lag horizons within the 240s prediction horizon (p $<$ 0.001, 3.51 $<$ Cohen's \textit{d}).



Auditory workload prediction horizons of 60s and 120s required 30s lag horizons for acceptable correlation, whereas the 240s prediction horizon required a 240s lag horizon. 
The multivariate auditory workload prediction Friedman tests identified significant differences between lag horizons for the 60s (${\chi}^2(3,21) = 35.57 $, p $<$ 0.001, \textit{W} = 0.44), 120s (${\chi}^2(3,20) = 18.06 $, p $<$ 0.001, \textit{W} = 0.24), and 240s (${\chi}^2(3,20) = 27.42 $, p $<$ 0.005, \textit{W} = 0.36) prediction horizons. 
The Wilcoxon signed-rank test indicated Spearman correlations increased significantly as the lag horizon increased in almost all cases (p $<$ 0.05, 0.79 $<$ Cohen's \textit{d}). The differences were not significant for 60s predictions between the 120s and 240s lag horizon, and between 60s and 120s lag horizons for the 120s and 240s prediction horizons.






\section{Discussion}

Predicting future workload will enable teleoperated systems to adjust to the human's available resources (i.e., the human's workload). Anticipating the human's internal resources is particularly critical for human-robot interactions with built-in delays, such as robots that may experience communication delays and dropouts. 
Existing workload predictions make predictions over very long prediction horizons (i.e., $>$ 1 hour), or very short predictions (i.e., $<$ 5s) \cite{yu2023towards, wei2023classification} that may be indistinguishable from na\"ive forecasts that use the last known workload estimate. 
Existing research also does not provide guidance on lag horizon selection for workload prediction. 
This paper's workload prediction models predict workload out to four minutes for the major supervisory tasks' workload components, with predictions validated against na\"ive forecasts. Minimum and maximum lag horizons are identified for the cognitive, visual, auditory and overall workload for each prediction horizon.

The last known workload estimate alone (i.e., na\"ive forecasts) was found to be insufficient for predicting the individual changes in workload 
(e.g., if visual or auditory workload will cause high overall workload). 
The univariate and multivariate workload prediction methods were found to have improved prediction accuracy and prediction horizons with longer lag horizons. 
Minimum lag horizon requirements varied between workload components, indicating optimization may be performed on an individual workload component basis. 
Notably, the cognitive workload broadly required the shortest lag horizons, and visual workload the longest.

The univariate method made predictions based on the historical trends within a specific workload component. 
Longer lag horizons improved the univariate workload predictions; however, as the prediction horizon increased, these benefits decreased. 
Notably, a maximum lag horizon was not found for the univariate predictions. 
Univariate workload predictions generally required 240s lag horizons to achieve acceptable results for predictions $>$ 60s, suggesting univariate workload predictions have difficulty making long-term workload predictions. 
However, the univariate predictions were able to make predictions longer than existing comparable methods (i.e., $>$ 30s \cite{grimaldi2024deep}), and for additional workload components (i.e., visual, auditory). 
These results suggest univariate workload predictions may be used when practitioners only have access to 
single workload component's estimates.

Multivariate predictions used the covariance between the workload components to predict future workload values. 
Prediction horizons up to 120s had significantly improved accuracy with lag horizons up to 120s, after which out-of-date historical information resulted in diminishing returns. 
Increased access to relevant historical information improved prediction accuracy until out-of-date information was included. 
The 240s prediction horizon accuracy only significantly increased with a 240s lag horizon. 
This 240s time length represented a point of particularly strong seasonality, where the model may use both trend and seasonality to make more accurate workload predictions. 
These results suggest two paradigms for lag horizon selection. Shorter prediction horizons may have optimal lag horizons set by the workload component autocorrelation length. 
Alternatively, longer prediction horizons rely on points of high seasonality to select lag horizons and extend prediction horizons.



Predicting future human workload is required to inform adaptations in time-delayed systems, such as teleoperation. These systems have built in lag, meaning workload estimates alone only enable reactive actions to address problems after they have occurred. 
The presented models predict workload for several minutes into the future using historical workload values, with additional workload components decreasing the required lag horizon. 
These approaches are expected to generalize to other domains, and act as guidelines for selecting lag horizons for future workload prediction approaches.

\section{Conclusion}

Two direct time series forecasting approaches predicted future workload states using historical workload estimates. 
Lag horizons were investigated to determine how much data is required for each prediction horizon, and when further historical information ceased to improve predictions. 
These findings suggest univariate predictions required longer lag horizons. Conversely, multivariate predictions' lag horizon may be set by the autocorrelation length or specific points of high seasonality. 
These workload predictions will enable teleoperated systems to anticipate the human's internal resources in scenarios with communication delays and dropouts, and improve performance through workload-aware adaptations.



\normalsize

\bibliographystyle{IEEEtran}
\bibliography{sample-base.bib}

\end{document}